# A Course Intelligence Platform for Higher Education: Lessons from AI-Assisted Course Evaluation

Bo Yuan[1*[0000-0003-2169-0007]], Jiazi Hu[2] and Haimei Zhao[3[0000-0003-1139-4183]]

[1] The University of Queensland, QLD 4072, Australia
[2] AI U Course, Kunming 650000, P.R. China
[3] The University of Sydney, NSW 2006, Australia
boyuan@ieee.org, jessica@aiucourse.com, h.zhao@sydney.edu.au

**Abstract.** The rapid adoption of generative AI has created new opportunities for teaching, learning, and quality assurance. Existing applications, however, remain largely student-facing, with comparatively limited attention to institution-level needs. This paper presents a course intelligence platform deployed across more than 100 universities and serving over 10,000 instructors in China. By linking competency requirements, knowledge structures, teaching activities, and assessment evidence, it establishes a shared foundation for knowledge organization, instructional design, learning assessment, and quality evaluation. The course evaluation module is examined as a representative institution-facing application of the platform, which integrates national evaluation standards, structured educational evidence, customized prompting strategies, and domain-adapted LLMs to generate quantitative scores and qualitative feedback. A case study involving 100 authentic university courses is conducted to evaluate its alignment with expert judgments and the interpretability of its outputs. Statistical analyses show substantial agreement between AI-generated assessments and expert ratings, while qualitative results highlight the credibility of the feedback. The findings further suggest that AI-assisted evaluation requires not only capable models but also structured domain knowledge and transparent criteria. In this context, human ratings should be treated as an informative reference rather than an error-free gold standard, and the objective is to achieve consistent, interpretable, and defensible judgments instead of merely replicating expert scores.

**Keywords:** Course Evaluation, Large Language Models, Higher Education, Quality Assurance, Educational Governance.

## 1 Introduction

The emergence of generative AI, particularly large language models (LLMs) [1], is reshaping higher education. Recent studies have demonstrated its potential across a wide range of activities, including tutoring, student assessment, personalized learning, and instructional content generation [2–4]. Accordingly, AI is increasingly viewed not merely as a collection of standalone learning tools but as a broader technical foundation for educational innovation, institutional governance, and quality enhancement at scale.

However, existing AI applications remain predominantly student-facing, with emphasis on assignment marking, feedback generation, adaptive learning, and tutoring [5–8]. While these applications have demonstrated considerable value for individual learning processes, institution-facing challenges critical to the sustainable development of higher education systems have received comparatively limited attention.

In practice, universities must routinely manage and improve a range of complex educational functions, such as programme design, curriculum development, and quality assurance. These activities generate and depend on growing volumes of data, including course syllabi, teaching plans, instructional resources, assessment records, learning outcomes, accreditation documents, and quality assurance reports. Together, these materials form a rich educational information ecosystem. However, the relevant information is often fragmented across multiple systems and stored predominantly in unstructured formats. As a result, many educational decisions still rely heavily on manual analysis, which limits both scalability and consistency.

To address these challenges, this paper introduces the concept of *course intelligence*: the transformation of heterogeneous course information into structured and actionable educational knowledge for both human decision making and AI-driven services. Building on this concept, we developed a course intelligence platform, which integrates curriculum repositories, knowledge graphs, and competency graphs to enable a range of AI services. It has been deployed across more than 100 universities in China and used by over 10,000 instructors. Continued adoption at this scale reflects both the platform's practical value and institutional acceptance in real-world settings.

Among these applications, course evaluation is one of the most resource-intensive and strategically important functions in quality assurance [9]. Large-scale reviews often require experts to examine extensive teaching materials and judge courses against standardized criteria. Such activities are widely recognized as time-consuming, difficult to scale, and susceptible to inconsistency [10]. This resource intensity often limits the frequency of formal course evaluations to, for example, once per academic year, delaying feedback and reducing the timeliness of quality improvement. Course evaluation therefore provides a realistic setting for examining whether generative AI can support more timely and scalable approaches to institutional educational governance.

Rather than presenting a benchmark-oriented comparison of LLMs [11], this paper examines their performance on a specific evaluation task. Based on 100 courses drawn from an authentic review process, we investigate the operational behavior, alignment with expert judgments, interpretability, and practical feasibility of AI-assisted course evaluation. These analyses help clarify whether AI-generated judgments can provide a credible and practically usable basis for human–AI collaboration in institutional quality assurance. The contributions of this work are threefold:

- We present a course intelligence framework that integrates heterogeneous course information to support multiple educational applications.
- We report a case study involving 100 university courses and examine the agreement between AI-generated evaluations and expert judgments.
- We summarize practical lessons from deploying generative AI in authentic institutional settings and discuss their implications for educational governance.

## 2 Course Intelligence Platform

### 2.1 Platform Overview

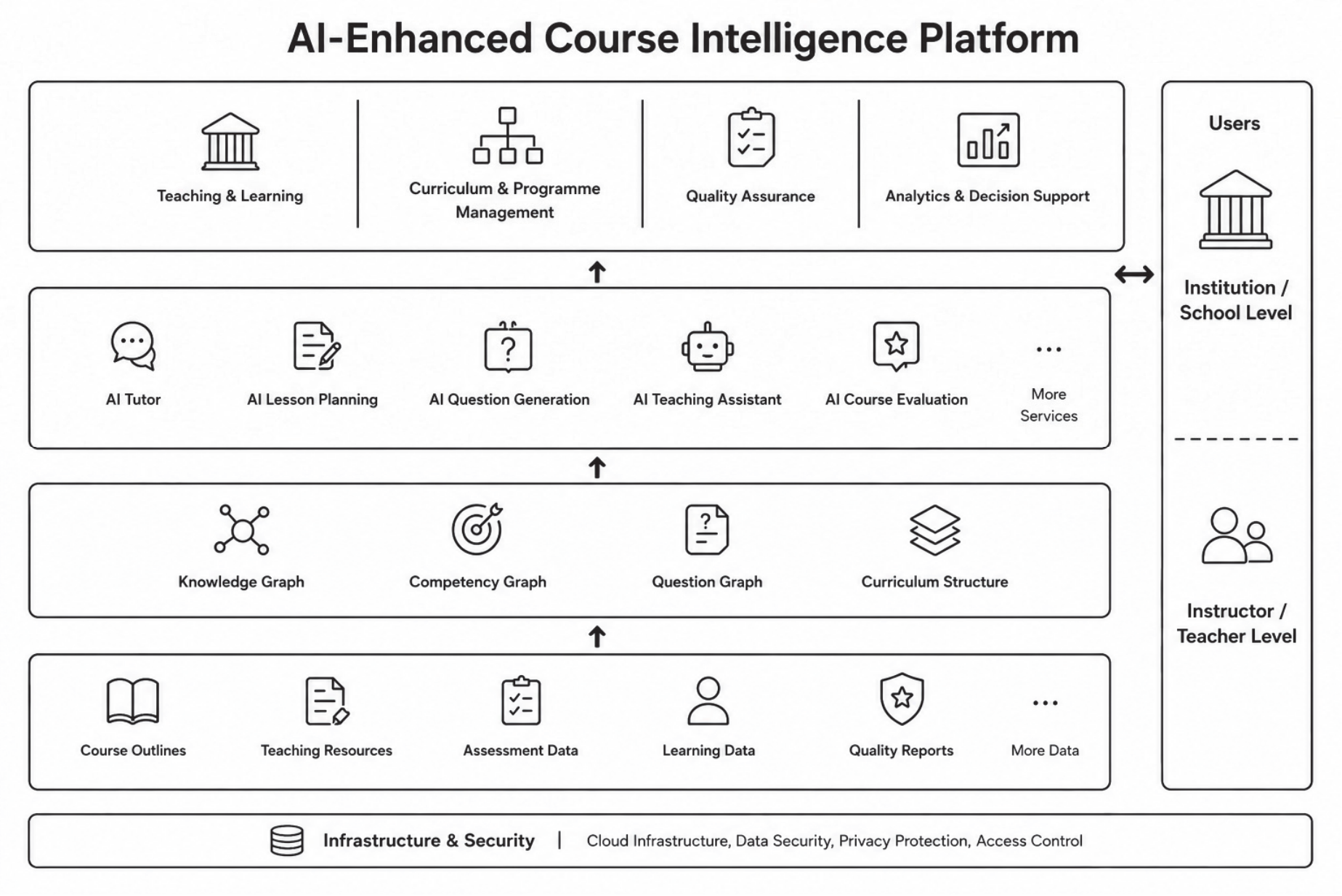

**Fig. 1.** Architecture of the proposed course intelligence platform. The system provides multiple AI-enabled services for teaching, learning, curriculum management, and quality assurance at both institutional and instructor levels.

The platform is organized as a layered architecture that connects course data, structured educational representations, AI services, and higher-level applications (Fig. 1). At the foundation, heterogeneous course-related data are collected from sources such as course outlines, teaching resources, assessment records, learning data, and quality reports. These data are transformed into structured representations, including knowledge graphs, competency graphs, question graphs, and curriculum structures, which provide a common information layer for downstream applications. Building on this foundation, the platform provides AI-enabled services such as tutoring, lesson planning, question generation, teaching assistance, and course evaluation. These services are further integrated into broader teaching, curriculum and programme management, quality assurance, and analytical workflows.

### 2.2 Course Intelligence Layer

The central concept underlying the platform is course intelligence. Traditionally, course-related resources are distributed across multiple documents, including syllabi, teaching plans, instructional materials, assessment records, and quality assurance reports. Although these materials collectively describe the key educational features of a course, they are often fragmented and difficult to analyze systematically. To address this challenge, the platform transforms heterogeneous course information into structured educational representations (Appendix A). These include:

- **Instructional Model**, describing learning objectives, teaching content, assessment strategies, and the overall organization of instruction;
- **Knowledge Graph**, representing domain knowledge structures and relationships among concepts covered in a course;
- **Competency Graph**, capturing the competencies, skills, and learning outcomes expected of students;
- **Question Graph**, linking assessment items to knowledge concepts and competency requirements for learning assessment, diagnosis, and evaluation.

Together, these structured representations establish explicit connections among competencies, knowledge, instruction, and assessment, creating a reusable knowledge foundation for curriculum analysis, learning diagnosis, educational evaluation, and a wide range of AI-assisted teaching and governance services.

### 2.3 AI Service Layer

Building on the course intelligence layer, the platform provides a range of services powered by LLMs and educational knowledge bases. Key services include:

- **AI Tutor**, offering personalized learning assistance, question answering and guidance for students;
- **Lesson Planning**, facilitating instructional design, lesson preparation, and the planning of teaching activities;
- **Question Generation**, generating assessment items aligned with knowledge, learning objectives, and competency requirements;
- **AI Teaching Assistant**, assisting instructors with grading, feedback generation, and instructional coordination;
- **Course Evaluation**, enabling systematic course review, evidence-based quality assurance, and continuous enhancement of course design and delivery.

### 2.4 Human–AI Collaboration in Practice

A key feature of the platform is its emphasis on human–AI collaboration. Rather than replacing educators, the system is designed to augment expert decision making through scalable analysis, structured feedback, and evidence-based recommendations. In this collaborative setting, AI performs large-scale information processing and preliminary analysis, while human experts retain responsibility for contextual interpretation, professional judgment, and final decisions. This division of roles allows AI-generated outputs to support, rather than determine, educational decisions. Within the platform, course evaluation represents a particularly important use case because it requires the systematic integration of diverse educational evidence, standardized criteria, and expert judgment to assess course quality and identify areas for improvement. Therefore, the course evaluation module was selected as a case study for investigating the real-world deployment of generative AI in institutional quality assurance.

## 3 AI-Assisted Course Evaluation

### 3.1 Motivation

Universities regularly conduct course review to assess curriculum quality, instructional effectiveness, assessment design, and continuous improvement efforts. In large-scale institutional review processes, experts are often required to examine extensive documentation and judge courses against standardized criteria. Traditional expert-based workflows, however, face several practical challenges. First, substantial numbers of courses may need to be assessed within limited timeframes, creating considerable workload pressure. Second, outcomes may be influenced by reviewer experience, disciplinary background, and scoring tendencies, making consistency difficult to maintain even under common evaluation criteria. To address these challenges, the AI-assisted course evaluation module is developed to provide scalable analysis, structured feedback, and evidence-based results. Its objective is not to replace expert judgment, but to improve the efficiency, consistency, and interpretability of course review.

### 3.2 Evaluation Criteria

The course evaluation process follows nationally published standards widely adopted in higher education quality assurance in China. The framework consists of multiple first-level and second-level indicators covering key dimensions such as course design, instructional team, teaching resources, assessment design, and teaching effectiveness (Table 1). By adopting an established framework, the AI system operates within authentic educational governance practices rather than relying on artificially constructed benchmark tasks. The overall performance is reported on a 100-point scale, with scores above 85 generally indicating excellent quality. In practice, however, the proportion of courses receiving this rating is often deliberately limited to preserve the discriminative value of the assessment and avoid score inflation.

**Table 1.** Major evaluation dimensions used in the course evaluation module.

| First-Level Indicators | Weight (%) | Second-Level Indicators |
|---|---|---|
| Course Design | 25 | Course Objectives, Course Content, Teaching Organization, Teaching Methods |
| Instructional Team | 20 | Professional Ethics, Team Structure, Team Quality and Competence |
| Teaching Resources | 20 | Resource Selection, Resource Development, Practical Teaching |
| Course Assessment | 10 | Assessment Content, Assessment Methods |
| Teaching Effectiveness | 25 | Student Development, Diverse Assessment, Continuous Improvement |

### 3.3 LLM-Based Evaluation Flow

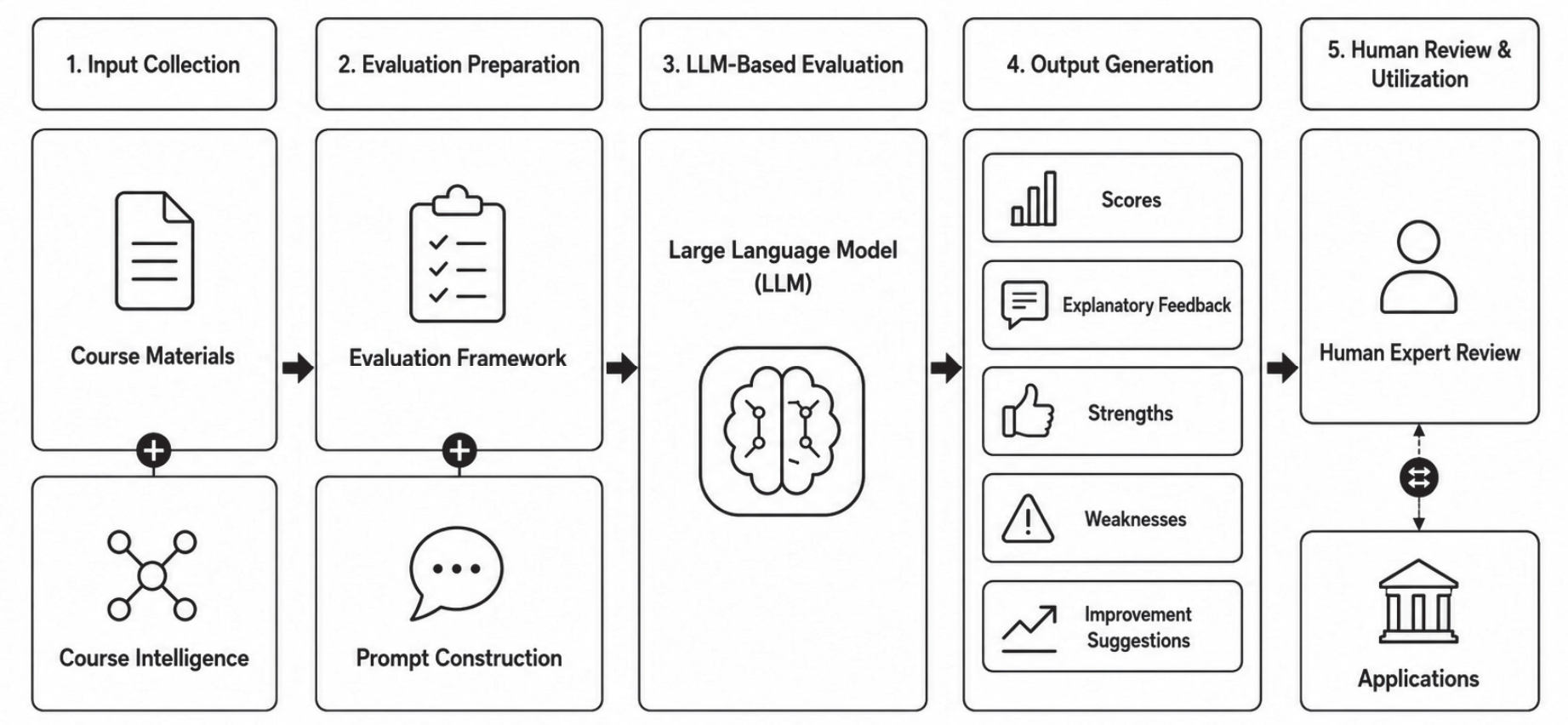

**Fig. 2.** Workflow of the course evaluation module. Educational evidence, course representations, and evaluation criteria are integrated to support LLM-based evaluation.

The evaluation workflow combines educational evidence, standardized criteria, prompt engineering, and LLMs to produce both quantitative and qualitative results (Fig. 2). Structured prompts incorporate the relevant evidence and criteria while specifying the evaluator role, scoring requirements, output format, and grounding constraints. To enhance reliability, the prompts are explicitly designed [12] to encourage evidence-based reasoning, meaningful score differentiation, and pedagogically grounded justification. An illustrative prompt template is presented in Appendix B. The outputs include an overall score, indicator-level scores, explanatory feedback, identified strengths and weaknesses, and actionable recommendations for improvement. Before practical use, all AI-generated evaluation results are reviewed and validated by human experts, who retain final decision-making authority.

### 3.4 Models and Configuration

Preliminary testing was conducted using several mainstream LLMs. Based on operational stability and output quality, two models were selected for further investigation: Kimi [13] and Llama [14]. To better align model behavior with expert practices, a domain-adapted version of Llama, referred to as Llama-UKP, was developed through fine-tuning on historical expert-rated examples from previous review activities [15]. Subsequent analysis therefore focused primarily on Kimi and Llama-UKP. The purpose of this study is not to rank competing models, but to examine the extent to which AI-generated assessments align with expert judgment in authentic evaluation workflows. Comparative experiments were included to illustrate how different model configurations affect scoring behavior rather than to benchmark the underlying capabilities of individual LLMs. Given the rapid evolution of LLMs, the results of this study are not intended to support general claims about the performance of any particular model.

## 4 Empirical Evaluation

### 4.1 Evaluation Procedure

To examine the applicability of AI-assisted course evaluation in real-world educational settings, a case study was conducted using data from a recent institutional review at a major university in China. As part of the review, course teams submitted comprehensive documentation, including teaching materials, assessment information, and supporting evidence for expert appraisal. These materials reflected authentic teaching practices and institutional quality assurance requirements, providing a realistic basis for assessing the operational characteristics of the AI-assisted approach.

A total of 100 courses were included in the study, spanning diverse disciplinary areas and course types, with substantial variation in instructional design and resource quality. Human expert judgments produced during the review process served as reference evaluations. For each course, the module analyzed the submitted materials according to the standardized framework and generated a structured report covering both quantitative assessment and qualitative feedback. The study did not seek to determine whether AI could replace experts. Instead, it examined the extent to which AI-generated assessments aligned with expert judgments and whether the generated outputs possessed sufficient interpretability and practical value for institutional course review.

### 4.2 Score Distribution and Scoring Behavior

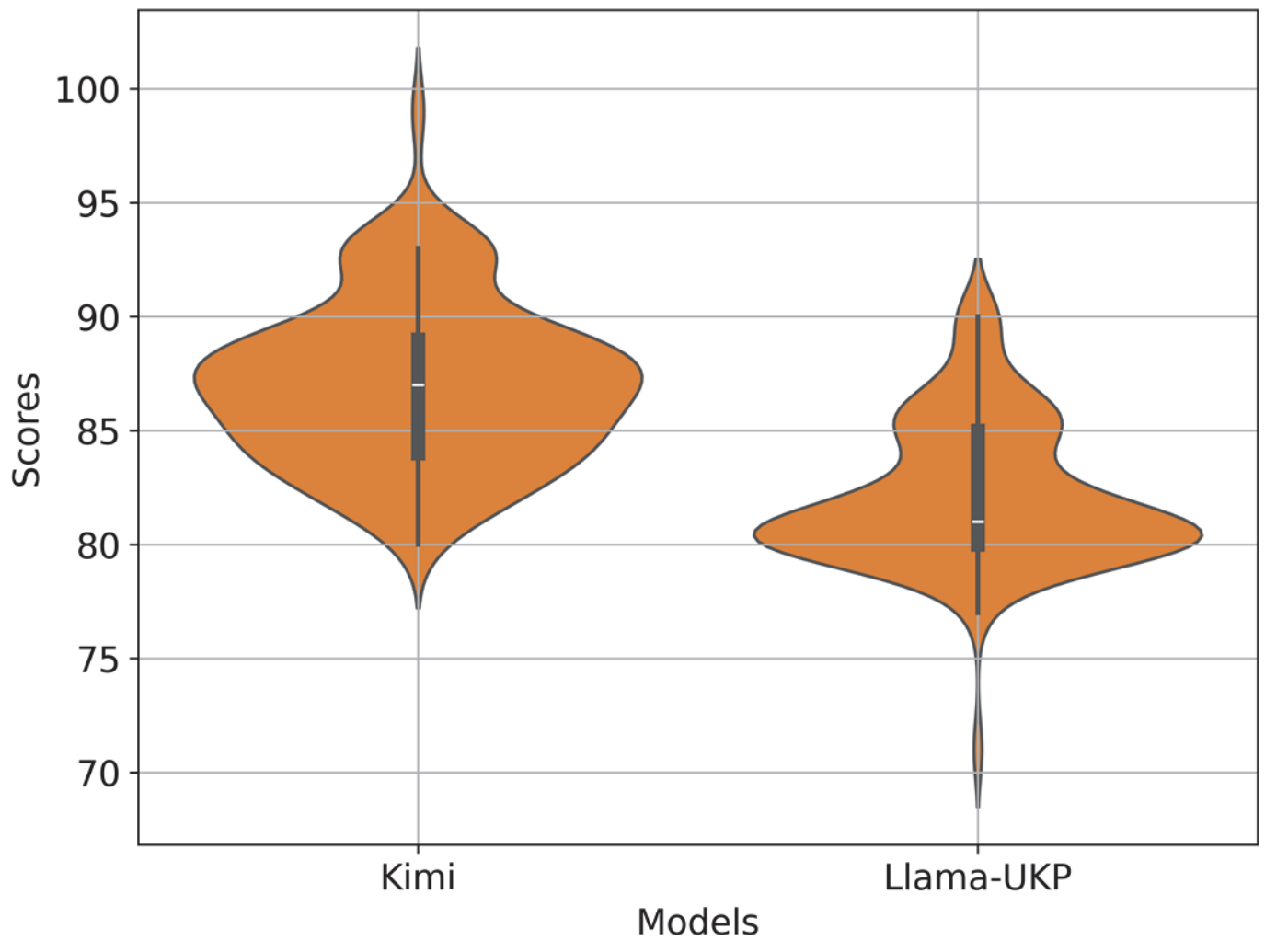

**Fig. 3.** Violin plots of course scores generated by Kimi and Llama-UKP across 100 courses. The two models exhibit distinct scoring distributions, with Llama-UKP producing a more concentrated pattern and Kimi showing a stronger tendency toward higher ratings.

Fig. 3 illustrates the overall score distributions generated by Kimi and Llama-UKP. Llama-UKP produced a more concentrated distribution, with the highest density around 81 points, whereas Kimi exhibited a broader pattern with a peak near 87 points. The difference is particularly evident in the proportion of courses classified as excellent: 27% of courses evaluated by Llama-UKP received scores of 85 or above, compared with 73% under Kimi. These results indicate a pronounced high-score tendency in the general-purpose model, while substantially fewer Llama-UKP scores exceeded the threshold. As the violin plots are based on kernel density estimation, the displayed density may extend slightly beyond the observed score ranges.

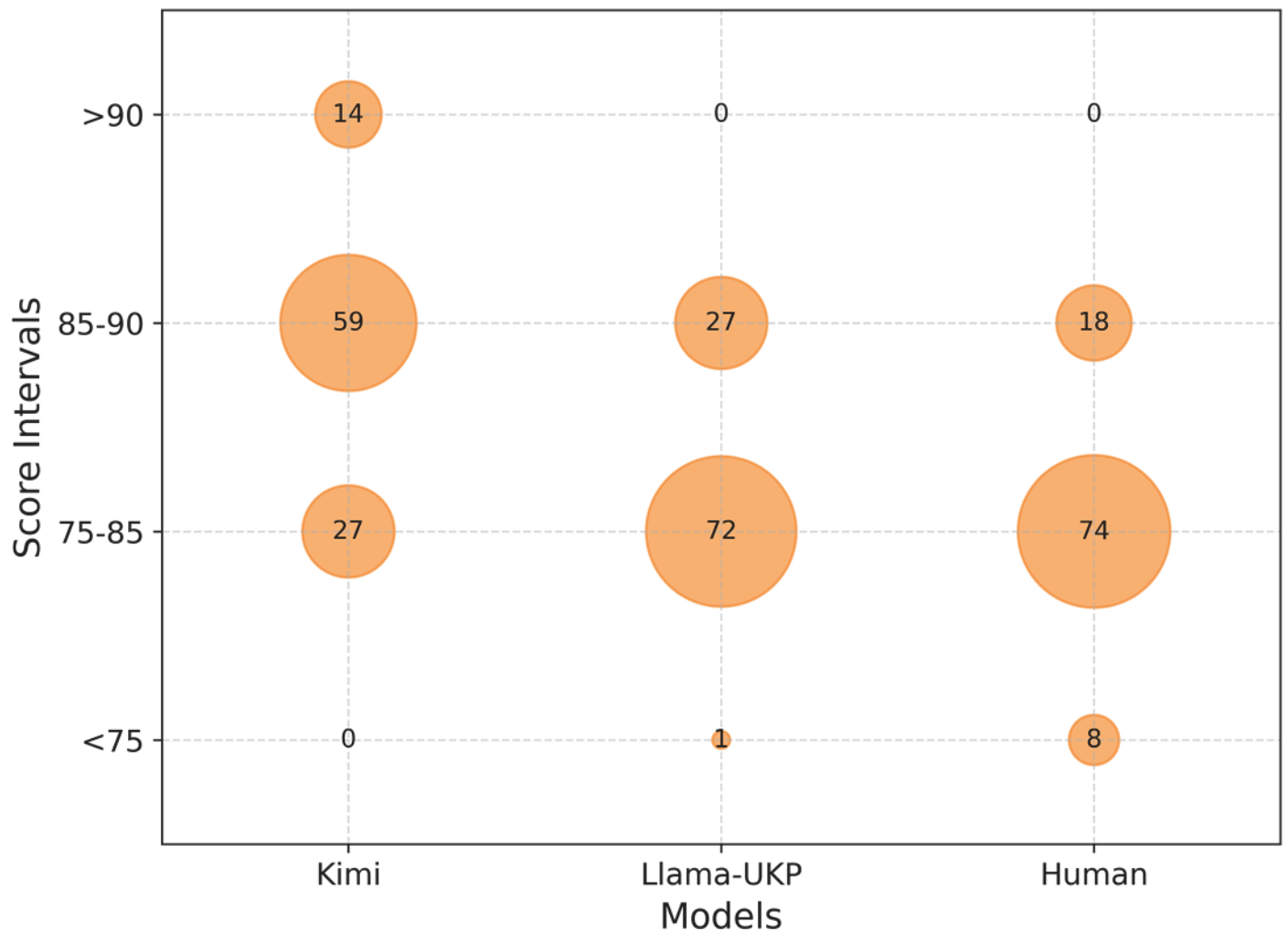

**Fig. 4.** Bubble plots comparing the score distributions produced by the two LLMs with human assessments across 100 courses. The domain-adapted Llama-UKP exhibits a scoring pattern that more closely resembles expert judgments.

Fig. 4 further compares the score distributions produced by the two models with human assessments. Kimi tended to assign higher scores and showed a scoring pattern that differed noticeably from that of human experts. By contrast, the distribution produced by Llama-UKP more closely resembled expert assessments in both its central range and overall differentiation. This comparison suggests that effective course evaluation depends not only on general language capability but also on alignment with institutional scoring practices and educational conventions. Domain adaptation therefore appears to improve alignment with real-world review standards.

### 4.3 Agreement with Human Experts

To assess the correspondence between AI-generated evaluations and expert judgments, several analyses were conducted. As shown in Table 2, Llama-UKP achieved a Pearson correlation of 0.754 and a Spearman correlation of 0.843 with human assessments, compared with 0.516 and 0.523 for Kimi, respectively. These results indicate stronger alignment between the domain-adapted model and expert judgments.

**Table 2.** Correlation between AI-generated evaluations and human assessments.

| Model | Pearson's *r* | Spearman's *ρ* |
|---|---|---|
| Llama-UKP | 0.754 | 0.843 |
| Kimi | 0.516 | 0.523 |

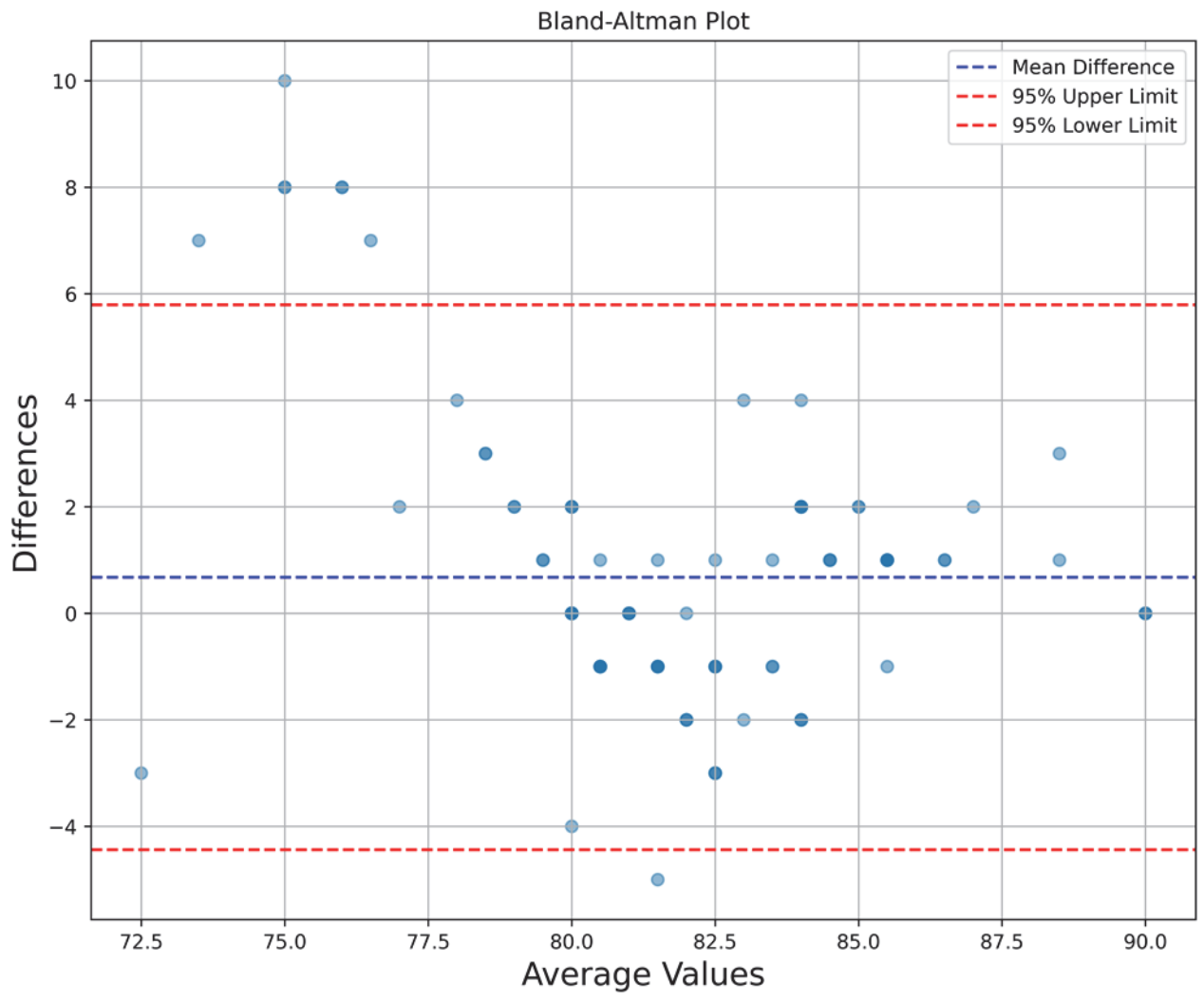

**Fig. 5.** Bland–Altman plot comparing scores generated by Llama-UKP with human ratings. Most observations fall within the limits of agreement, suggesting limited systematic bias and generally acceptable consistency between the two sets of scores.

Bland–Altman analysis [16] was also conducted to assess the agreement between scores generated by Llama-UKP and those assigned by human experts. As shown in Fig. 5, most score differences fell within the 95% limits of agreement, with little evidence of systematic bias across the rating range. Combined with the correlation results, these findings suggest that the discrepancies were more consistent with differences in score calibration than with fundamentally different assessment patterns.

To provide a more rigorous reliability analysis, intraclass correlation coefficients (ICCs) [17] were calculated. ICC evaluates the extent to which ratings agree within a common measurement framework, providing a basis for assessing their comparability across raters. In Table 3, the ICC(C,1) value of 0.734 indicates that Llama-UKP largely preserved the discriminative structure observed in expert evaluations, maintaining a consistent relative ordering of course quality. The ICC(A,1) value of 0.723 further suggests substantial agreement on the original scoring scale. Notably, the small difference between the two coefficients ($\Delta$=0.011) indicates only limited systematic differences in score calibration relative to expert ratings.

**Table 3.** ICC-based agreement between Llama-UKP and human assessments.

| Metric | Value | Interpretation |
|---|---|---|
| ICC(A, 1) | 0.723 | Absolute agreement |
| ICC(C, 1) | 0.734 | Consistency |
| Δ | 0.011 | Difference |

### 4.4 Qualitative Analysis

**Table 4.** Examples of AI-generated scores and feedback for the course objective indicator.

| Score | Feedback |
|---|---|
| 4.7/5 | The course objectives align well with the goals of the statistics major, emphasizing mathematical foundations, experimental skills, and regression analysis. Through diverse instructional activities, the course fosters independent thinking and innovation, enhances students' data analysis skills and understanding of statistical concepts, and supports holistic student development. |
| 3.8/5 | The course objectives closely align with the training goals of the computer science major, emphasizing the systematic development of programming skills. However, the connection to career development requires further clarification, and greater emphasis on higher-order thinking and innovation skills is recommended. Enhancing these aspects would make the objectives more specific and improve teaching effectiveness. |

Table 4 presents examples of the scores and feedback generated by Llama-UKP for two selected courses under the course objective indicator. The model generally linked evaluation scores to specific pedagogical evidence drawn from the submitted materials. Higher-scored courses tended to demonstrate clearer alignment with disciplinary goals, stronger emphasis on higher-order thinking, and more comprehensive competency development. Lower-scored courses, by contrast, more often received comments concerning insufficient specificity, innovation, or practical relevance. These examples illustrate how such outputs can provide well-grounded justifications by linking judgments to course documentation, thereby supporting trust in AI-assisted evaluation.

## 5 Lessons Learned from Real-World Deployment

### 5.1 Domain Knowledge vs. Generic Model Capability

The deployment experience suggests that the effectiveness of AI-assisted evaluation depends not only on the capability of the underlying LLMs but also on the availability of structured knowledge and clearly defined evaluation frameworks. Generic LLMs may generate fluent responses, but meaningful and consistent judgments require contextual information that defines what should be assessed and how decisions should be made. The findings therefore underscore the importance of embedding AI within institutional knowledge structures, rubrics, and governance mechanisms that shape and constrain its analytical outputs. Educational standards, assessment criteria, and institutional policies remain the primary sources of evaluative authority, while AI should be viewed as an analytical mechanism operating within established structures, rather than as an independent source of authority [18].

### 5.2 Relative Ranking vs. Absolute Calibration

Another important finding concerns the distinction between relative consistency and absolute agreement. The analysis showed that AI-generated assessments were effective at distinguishing stronger from weaker courses and preserving ranking patterns similar to those of human evaluators, while exact numerical calibration remained challenging. Importantly, perfect agreement with individual expert scores should not be regarded as the ultimate criterion for AI-assisted evaluation. Human judgments are themselves subject to subjective and contextual variation [19] and therefore do not constitute an error-free gold standard. Expert ratings are better treated as a reference for assessing whether AI-generated results are broadly consistent with established evaluative practices. In practice, preserving meaningful differentiation among courses and producing defensible, evidence-based judgments may be more valuable than matching the exact numerical scores assigned by human reviewers. From this perspective, validity depends less on correspondence with any single expert rating than on consistent alignment with accepted evaluation principles.

### 5.3 Interpretability and Institutional Adoption

Trust and transparency emerged during deployment as essential conditions for institutional adoption [20]. As quality assurance involves consequential judgments, acceptance depends not only on scoring performance, but also on whether AI-generated decisions are transparent, traceable, and subject to human review. Interpretability in this context should not be equated with the production of plausible rationales. Instead, explanations should be anchored in identifiable evidence from the reviewed materials and explicitly connected to the relevant criteria and resulting scores. The ability to trace judgments to supporting evidence, identify strengths and weaknesses, and generate actionable recommendations is critical for accountability and professional oversight. Interpretability therefore constitutes a core requirement for trustworthy AI-mediated judgment and effective human–AI collaboration in quality assurance.

### 5.4 The Effect of Fine-Tuning

To examine the effect of fine-tuning, this study compared Llama-UKP with the original Llama. The original model exhibited greater score dispersion, including higher variance, a wider overall range, and lower-tail values extending to approximately 60. Although the two models showed similar medians, Llama-UKP produced a substantially more concentrated distribution with fewer extreme ratings. Human evaluators are well known to exhibit central-tendency and range-restriction effects, often using a limited portion of the available scale and avoiding extreme categories [21]. As Llama-UKP was fine-tuned on expert-rated data, its narrower score distribution suggests that fine-tuning captured not only the scoring levels reflected in those examples but also aspects of the underlying rating behavior of human experts. In this sense, it may act as a form of distributional calibration, bringing model-generated judgments closer to both the scoring levels and statistical patterns observed in expert assessments.

## 6 Conclusion

This paper presented a course intelligence platform for curriculum development, teaching innovation, and quality assurance in higher education. By integrating heterogeneous educational resources, structured course representations, and LLMs, the platform provides a shared basis for diverse AI-assisted services. As a representative application of the platform, the course evaluation module was examined through a case study involving 100 university courses. The results showed that AI-generated evaluations can exhibit substantial agreement with expert judgments, while providing structured, interpretable, and pedagogically meaningful feedback. The deployment experience further underscored the importance of domain knowledge, transparent evaluation criteria, and evidence-grounded explanations in ensuring that AI-assisted assessment remains consistent with institutional requirements.

Beyond the continued expansion of functional capabilities [22], future work should place greater emphasis on the human and institutional dimensions of AI-assisted evaluation. One of the most significant challenges is not technical feasibility itself, but whether AI-generated judgments can gain trust and legitimacy among administrators and educators and become part of established quality assurance practices. Addressing this challenge requires a deeper understanding of how human and AI judgments should be combined, how evaluative authority should be distributed, and how transparency, accountability, and professional autonomy can be preserved in AI-mediated decision making. More broadly, the long-term significance of such systems lies not simply in automating existing review tasks, but in reshaping how educational judgments are produced, negotiated, and governed across the higher education landscape. The central question, therefore, is not whether AI can participate in evaluation, but how universities can establish forms of human–AI collaboration that are both educationally credible and institutionally legitimate.

**Acknowledgments.** The authors acknowledge the use of AI-assisted writing tools in preparing this manuscript. The authors retain full responsibility for the research design, experimental workflow, interpretation of the results, and final content of the paper.

**Disclosure of Interests.** The authors declare no competing interests.

# Appendices

## A. Course Intelligence Representation

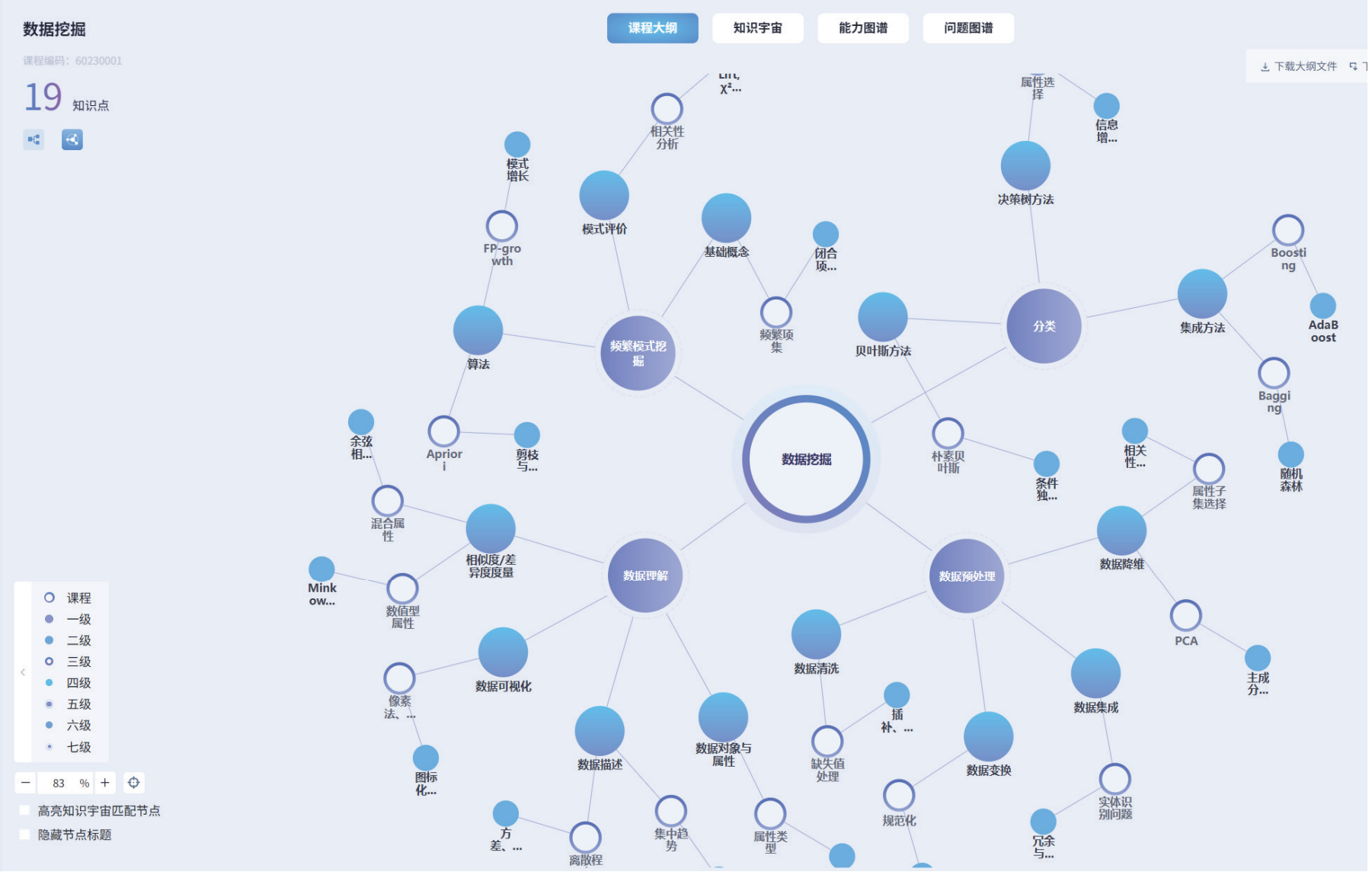

**Fig. A.** Example knowledge graph automatically generated from the course outline of the Data Mining course. The graph captures hierarchical relationships among major topics and concepts, providing a structured representation of course knowledge.

A key function of the platform is to transform heterogeneous course materials into structured educational representations that can be directly utilized by downstream AI services. In conventional course delivery, relationships among knowledge concepts, learning objectives, and assessment activities are often implicit, making it difficult to relate student performance to specific concepts and competency requirements. Fig. A illustrates how such relationships can be made explicit through a knowledge graph generated from the Data Mining course outline. By modeling these connections systematically, the platform establishes a reusable knowledge foundation for curriculum analysis, learning diagnosis, personalized learning, educational evaluation, and a wide range of teaching and governance applications.

## B. Prompt Template

The course evaluation module employs a structured prompting strategy to align model behavior with institutional review practices. The prompts explicitly define evaluator roles, assessment criteria, evidence requirements, scoring expectations, and output formats. In addition to providing procedural instructions, the prompting framework translates key elements of the evaluation process into an explicit reasoning structure for the model, specifying what evidence should be considered, how judgments should be formed, and how scores should be justified. This design helps constrain overly generic or unsupported judgments by requiring the model to connect its conclusions to predefined criteria and identifiable course evidence. The following template summarizes the main components used in the study.

*1) Role Definition.*
The model is instructed to act as an experienced higher-education course reviewer responsible for assessing course quality according to established standards and institutional expectations.

*2) Evaluation Context*
The model receives a collection of course-related materials, including:

- course objectives;
- teaching plans;
- instructional resources;
- assessment information;
- supporting evidence;
- evaluation rubric.

*3) Evaluation Requirements*
The model is instructed to:

- assess the course according to the predefined rubric;
- assign scores to individual indicators and the overall course quality;
- provide evidence-based justifications for assigned scores;
- avoid unsupported positive judgments;
- maintain meaningful differentiation across courses;
- ensure consistency between quantitative ratings and qualitative comments.

*4) Output Format*
For each course, the model generates:

- an overall score;
- indicator-level scores;
- identified strengths;
- identified weaknesses;
- evidence-based explanations;
- actionable recommendations for improvement.